\documentclass[11pt,a4paper]{article}
\usepackage[hyperref]{emnlp2020}
\usepackage{times}
\usepackage{latexsym}

\usepackage{tabularx}
\usepackage{xcolor}
\usepackage{breqn}
\usepackage{amssymb}
\usepackage{pifont}

\newcommand{\cmark}{\ding{51}}%
\newcommand{\xmark}{\ding{55}}%

\usepackage{graphicx, wrapfig, subcaption, setspace, booktabs}

\usepackage{microtype}
\usepackage{enumitem}
\aclfinalcopy 
\def\aclpaperid{***} 

\title{Unsupervised Evaluation for Question Answering with Transformers}

\author{Lukas Muttenthaler$^{\dagger \ddagger}$~Isabelle Augenstein$^{\dagger}$~Johannes Bjerva$^{\dagger \odot}$ \\
  $^{\dagger}$ Dept. of Computer Science, University of Copenhagen \\
  $^{\ddagger}$ Max-Planck-Institute for Human Cognitive and Brain Sciences, Leipzig \\
  $^{\odot}$ Dept. of Computer Science, Aalborg University \\
  \texttt{muttenthaler@cbs.mpg.de, augenstein@di.ku.dk, jbjerva@cs.aau.dk}}

\date{}

\begin{document}
\maketitle
\begin{abstract}
It is challenging to automatically evaluate the answer of a QA model at inference time.
Although many models provide confidence scores, and simple heuristics can go a long way towards indicating answer correctness, such measures are heavily dataset-dependent and are unlikely to generalise.
In this work, we begin by investigating the hidden representations of questions, answers, and contexts in transformer-based QA architectures.
We observe a consistent pattern in the answer representations, which we show can be used to automatically evaluate whether or not a predicted answer span is correct.
Our method does not require any labelled data and outperforms strong heuristic baselines, across 2 datasets and 7 domains.
We are able to predict whether or not a model's answer is correct with 91.37\% accuracy on SQuAD, and 80.7\% accuracy on SubjQA.
We expect that this method will have broad applications, e.g., in semi-automatic development of QA datasets.

\end{abstract}

\section{Introduction}
\label{section:intro}

Evaluation of a QA model usually requires human-annotated answer spans to compare a model's output with.
At inference time, however, it is hard to automatically estimate whether an extracted answer span is correct.
While many models can provide confidence scores, and other heuristics might be used to deduce whether a prediction is correct, such measures are heavily dataset-dependent and are not likely to generalise.
Hence, given a new domain, a costly procedure of human annotation needs to be initiated in order to provide an estimate of the model's accuracy.
However, this approach naturally does not scale well to new unlabelled sequences. 

\begin{figure}[t]
    \centering
    \includegraphics[width=\columnwidth]{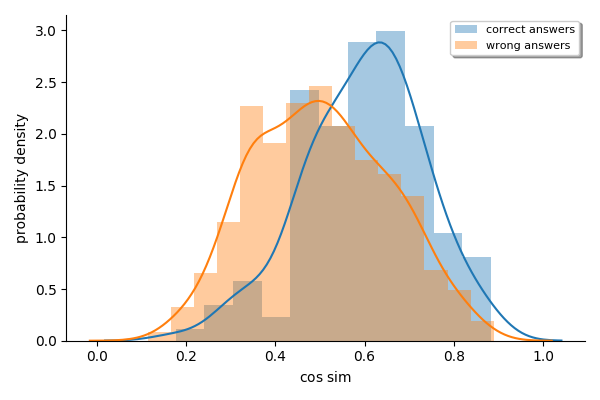}
    \caption{Probability Density Function of the cosine similarities among tokens w.r.t. the true answer span in SQuAD. Correct answer predictions (blue) tend to have higher cosine similarities than wrong answer predictions (orange).}
    \label{fig:squad_pdf}
\end{figure}

\begin{figure}
    \centering
    \includegraphics[width=\columnwidth]{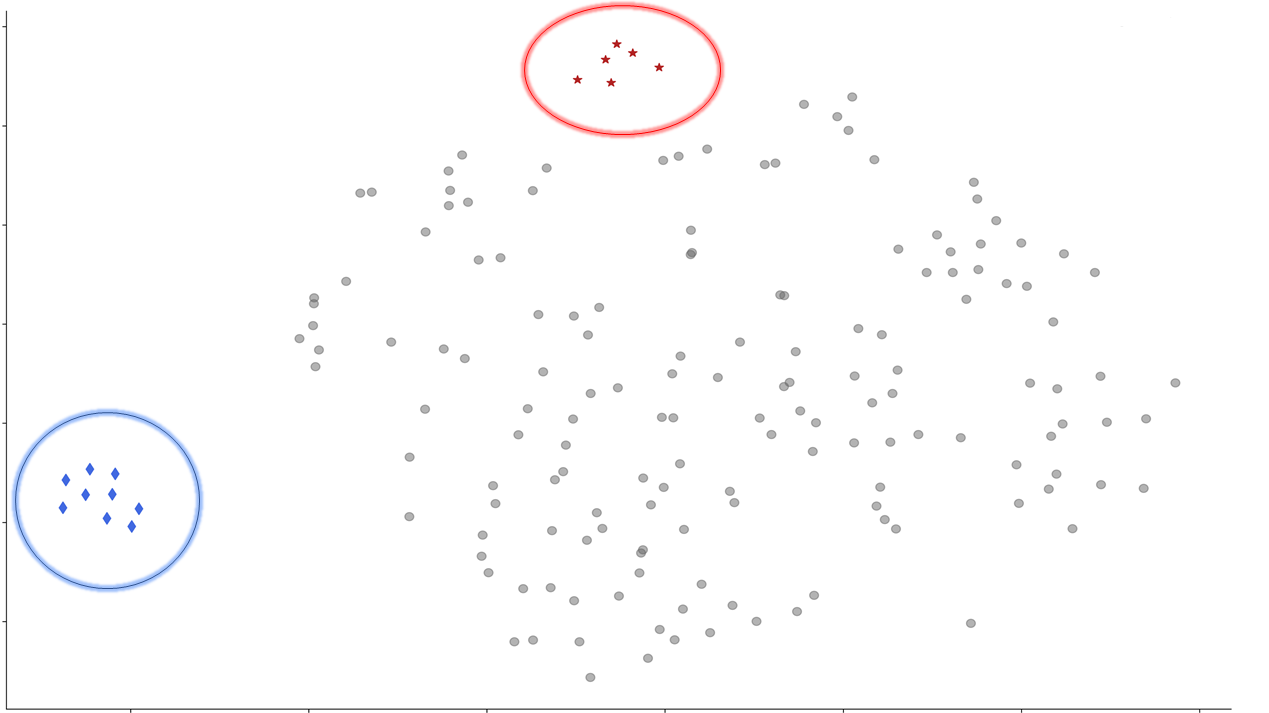}
    \caption{Hidden representations of answers are clustered separately from the remaining context for correct answer span predictions. This clustering is obtained by applying PCA followed by t-SNE (to save computational time), projecting the hidden representations for each token in a randomly chosen input sequence into $\mathbb{R}^{2}$. Blue diamonds: \textcolor{blue}{question}. Red stars: \textcolor{red}{answer}. Grey dots: \textcolor{gray}{context}.}
    \label{fig:word-cloud}
\end{figure}

In this work, we investigate Transformer-based QA models.
We hypothesise that hidden representations of later layers in such models contain information related to correctness of answers.
Indeed, we observe a consistent pattern of closely clustered answer token representations in the top three layers, whenever BERT correctly predicts an answer span (see Figure~\ref{fig:word-cloud}). 
Conversely, both true and predicted answer spans are clustered together with the remainder of the context, when an answer prediction is erroneous.
With clustering, we refer to the transformation of high-dimensional token representations into 2-dimensional vector space through the employment of PCA \cite{pca_2014}, followed by t-SNE \cite{vanDerMaaten2008}.
We furthermore see that correctly identified answer spans show a high mean cosine similarity across final layers (Figure~\ref{fig:squad_pdf}).
Before computing the cosine similarity between token representations, we apply PCA to remove noise, and preserve $95\%$ of the variance in the low-rank, orthogonal representation (see~\ref{method:answer_vec_agreements} for detailed information).

We demonstrate how this insight can be used to predict whether or not a prediction from a Transformer-based QA model is correct.
We evaluate our method across two distinct QA datasets in English, covering 7 domains.
We observe that the pattern can be used for automatic evaluation in both SQuAD v2.0 \citep{squad_unanswerable} and SubjQA \citep{subjqa2020}, a recently released dataset containing subjective questions and answers across several domains.
We show that we can evaluate such models without any labelled test data, with an accuracy of 91.37\% on SQuAD, and 80.7\% accuracy on the more challenging and diverse SubjQA.

\paragraph{Contributions} 
(i) We investigate how a Transformer-based model encodes correct and incorrect answer spans in its hidden representations;
(ii) We propose a method to leverage the information contained in these representations to predict whether a given answer span prediction is correct or not;
(iii) We demonstrate that a combination of our method with simple heuristics 
yields near-perfect predictions of answer correctness.

\section{Method}
\label{section:method}
\subsection{Data}
We experiment on two English-language QA datasets: SQuAD v2.0 \citep{squad_unanswerable} and SubjQA \citep{subjqa2020}.
Since SQuAD v2.0 exclusively contains objective questions that belong to a single domain, \texttt{Wikipedia}, we contrast this with the more diverse SubjQA.
SubjQA is a recently developed span-selection QA dataset that mainly consists of questions whose answer involves subjective opinions \cite{subjqa2020}.
Answer spans are extracted from review paragraphs that correspond to six different domains, namely \texttt{books}, \texttt{electronics}, \texttt{groceries}, \texttt{movies}, \texttt{restaurants}, \texttt{tripadvisor}.

\subsection{Experimental Setup}
For each of our implemented QA models, we use a pre-trained DistilBERT Transformer \cite{sanh2019distilbert} with one fully-connected output layer on top.\footnote{\url{https://huggingface.co/transformers/}}
Compared to BERT \cite{devlin2018bert}, with 12 layers in the base model, DistilBERT only contains 6 Transformer layers, without showing a statistically significant deterioration in performance on a variety of NLP downstream tasks \cite{glue2018,squad_unanswerable,sanh2019distilbert}.


We fine-tune BERT on either SQuAD v2.0 \cite{squad_unanswerable} or SubjQA \cite{subjqa2020} before investigating the hidden representations.
Since we analyse the similarity of hidden representations across answer span tokens, we only fine-tune BERT on answerable questions.
Unanswerable questions correspond to BERT's special \texttt{[CLS]} token. Therefore, a similarity analysis of hidden representations is not carried out for these.

\subsection{Answers are Separate from the Context} 
In order to investigate if any patterns are visible in the hidden representations, we project them into $\mathbb{R}^{2}$ via PCA \cite{pca_2014} and t-SNE \cite{vanDerMaaten2008} at each Transformer layer.
This layer-wise analysis reveals how the model clusters tokens in latent space at each stage of the model.
Figure~\ref{fig:word-cloud} shows this for every token in a randomly chosen sentence pair, for which the model correctly answered the questions.
Interestingly, the model clusters both the question and the answer separately from the context. We observe the same pattern with the standard BERT model, which is in line with one recent study \cite{bert_qa_layerwise}.
It is this pattern which we seek to investigate further.

\subsection{Answer Vector Agreements}
\label{method:answer_vec_agreements}

As depicted in Figure~\ref{fig:word-cloud}, the model's hidden representations for each token in the answer span are clustered more closely in vector space for correct compared to wrong answer span predictions.
This is particularly visible in the final three layers of the model, where high-level rather than low-level linguistic features are represented.
To verify this observation quantitatively, we compute the average cosine similarities among all hidden representations for each token in the answer span, whenever the correct answer contains more than a single token.
Hence, the following analysis was conducted exclusively for answerable questions since the correct answer span for unanswerable questions corresponds to the special \texttt{[CLS]} token.

Before this computation, we remove all feature representations corresponding to the special \texttt{[PAD]} token and transform the matrix of hidden representations $\mathbf{H}_{i} \in \mathbb{R}^{T \times D}$\footnote{$T$ is equal to the number of tokens in a sentence pair $(\mathbf{q, c})_{i}$ without appended \texttt{[PAD]} tokens and $D$ = $768$ which is the model's hidden size in each layer.} for each sentence pair sequence $(\mathbf{q, c})_{i}$ into a lower-dimensional space to remove noise and exclusively keep those principal components that explain the most variance among the feature representations.
In so doing, we use PCA \cite{pca_2014} and retain 95\% of the hidden representations' variance.
This yields a matrix of transformed hidden representations $\tilde {\mathbf{H}}_{i} \in \mathbb{R}^{T \times P}$ , for each sentence pair $(\mathbf{q, c})_{i}$.
From the transformed matrix of hidden representations, we extract the matrix of hidden representations corresponding to answer span tokens $\tilde {\mathbf{H}}_{a(i)} \in \mathbb{R}^{T_{a} \times P}$ to compute the average cosine similarity solely across all answer vectors. 

\subsection{Average Cosine Similarity}
\label{method:cos_ha}

The average cosine similarity among the rows of the answer representation matrix $\tilde {\mathbf{H}}_{a(i)} \in \mathbb{R}^{T_{a} \times P}$ is computed as follows,

\begin{footnotesize}
\begin{equation}
    \cos_{\tilde H(a)_i} = 2 \frac{\sum_{j}^{T_{a}} \sum_{k (k > j)}^{T_{a}} \cos(H_{a(i)}^{{j}^{T}}, H_{a(i)}^{{k}^{T}}) \in \mathbb{R}^{P}}{T_{a}T_{a}-T_{a}},
\label{equation:average_cosine}
\end{equation}
\end{footnotesize}

where the cosine similarity between two non-zero vectors $\mathbf{u}$ and $\mathbf{v}$ is defined as,

\begin{footnotesize}
\begin{equation}
    \cos (\mathbf{u}, \mathbf{v})=\frac{\mathbf{u} \cdot \mathbf{v}}{\|\mathbf{u}\|\|\mathbf{v}\|}=\frac{\sum_{i=1}^{n} u_{i} v_{i}}{\sqrt{\sum_{i=1}^{n} u_{i}^{2}} \sqrt{\sum_{i=1}^{n} v_{i}^{2}}}
\label{equation:cosine_sim}
\end{equation}
\end{footnotesize}

Since the cosine similarity is a symmetric metric we can compute the sum exclusively over the upper triangular of the similarity matrix (i.e., $\forall k > j$), thus saving computational time. The computation from Equation~\ref{equation:average_cosine} is performed for the two sets of correct and erroneous answer span predictions separately to inspect potential differences between the two w.r.t. their average cosine similarities. This was done at each Transformer layer $l \in L$, where $L$ = $6$, to examine shifts in the cosine similarity distributions across space.

\subsection{Probability Distributions in Correct and Erroneous Answers}
Based on having observed this pattern, we investigate how it extends to correct and incorrect answer span predictions.
Figure~\ref{fig:squad_pdf} shows that the probability to observe a high internal cosine similarity, $\cos_{\tilde H(a)_i}$, in later layers is significantly higher for correct compared to erroneous answer span predictions.

We can formalise the pattern by investigating the cumulative distribution function of the representations (CDF).
The probability $p_{(cdf)}^{l}$ that an observed $\cos_{\tilde H(a)_i}$ at Transformer layer $l$ lies in-between two cosine values can be obtained through the following interpolation, 

\begin{dmath}
 p_{(cdf)}^{l} = {P(\cos_{\mathbf{\tilde H}(a)}^{l} \leq \cos_{\tilde H(a)_i}^{l} + \: \Delta)} \\ - {P(\cos_{ \mathbf{\tilde H}(a)}^{l} \leq \cos_{\tilde H(a)_i}^{l} - \: \Delta)},
\end{dmath}

where $\cos_{\mathbf{\tilde H}(a)}^{l}$ denotes the train distribution of $\cos_{\tilde H(a)_{(i \: \in \: N)}}^{l}$ w.r.t. either correct or erroneous answer span predictions, and $\Delta$ is a hyperparameter that refers to the boundaries of the CDF interval.
$\Delta$ is set to $.1$ for all experiments, and 
$p_{(cdf)}^{l}$ is computed $\forall \: l \in L$.
\begin{table*}[t]
\centering
\begin{footnotesize}
    \begin{tabularx}{\linewidth} {@{}l||XX|XX|XX|Xr@{}}
    \toprule
    \textsc{Layer $\setminus$ Source}&\multicolumn{4}{c}{\textsc{SQuAD}}&\multicolumn{4}{c}{\textsc{SubjQA}}\\
    \midrule
    & \multicolumn{2}{c}{\textsc{Dev}} & \multicolumn{2}{c}{\textsc{Test}} & \multicolumn{2}{c}{\textsc{Dev}} & \multicolumn{2}{c}{\textsc{Test}} \\ 
    \midrule
    & $p$-value & diff. & $p$-value & diff. & $p$-value & diff. & $p$-value & diff. \\ 
    \midrule
    \textsc{layer} $1$ &  $.121$ &  $+.037$ & $.312$  & $+.031$ &  $.012$ &     $-.028$  & $\mathbf{.000}$*** &     $-.034$   \\
    \textsc{layer} $2$ &  $.069$  &  $+.040$  & $.256$  & $+.021$   & $.020$ & $-.036$  & $.001$** & $-.031$   \\
    \textsc{layer} $3$ &  $.007$**  &  $+.054$  & $.419$  & $+.028$ & $.185$ & $-.027$  & $.232$ & $-.020$   \\
    \textsc{layer} $4$ &  $.002$**  &   $+.061$ & $.422$  & $+.023$  &   $\mathbf{.000}$*** & $+.089$  & $\mathbf{.000}$*** &      $+.109$    \\
    \textsc{layer} $5$ &  $\mathbf{.000}$*** &  $+.151$ &  $\mathbf{.000}$*** &  $+.094$  &  $\mathbf{.000}$*** &  $+.115$  & $\mathbf{.000}$*** &      $+.133$   \\
    \textsc{layer} $6$ &  $\mathbf{.000}$***  &  $+.157$  & $\mathbf{.000}$*** & $+.095$  &  $\mathbf{.000}$*** &  $+.116$  & $\mathbf{.000}$***  & $+.129$    \\
    \bottomrule
    \end{tabularx}
\end{footnotesize}
\caption{Differences between correct and erroneous answer span predictions with respect to $\cos_{\tilde H(a)}$ (see Equation~\ref{equation:average_cosine}) at every Transformer layer. \textit{p}-values refer to statistically significant differences according to Bonferroni corrected independent \textit{t}-tests ($p < .05 =$*, $p < .01 =$ **, $p < .001 =$ ***, $p = .000$ $\leq 1e-4$). The difference in mean cosine similarities between prediction sets is captured by the diff. column. $+$ indicates higher $\cos_{\tilde H(a)}$ values for correct answer span predictions. Highly statistically significant differences (***) are marked in bold face.}
\label{tab:statistical_analyses}
\end{table*}

We compare the distribution $\cos_{\mathbf{\tilde H}(a)}^{l}$ corresponding to correct and erroneous answer span predictions respectively against each other $\forall \: l \in L$. 
We apply independent $t$-tests, and adjust $p$-values post-hoc using Bonferroni correction to counteract the multiple comparisons problem.
Analyses are performed for both development and test sets.

Table~\ref{tab:statistical_analyses} shows that $\mu$ with respect to $\cos_{\mathbf{\tilde H}(a)}^{l}$ is significantly higher ($p \leq 1\mathrm{e}{-4}$) for correct compared to erroneous answer span predictions for Transformer layers 5 and 6 across datasets.
Apart from SQuAD's test set, the same observation holds for Transformer layer 4.
This is in line with both boxplots, CDFs and PDFs, and indicates that an incorrect predictions starts being erroneous at layer 4.
This information can be conveniently leveraged for downstream applications, which is what we show in the following section by applying it to the evaluation of QA.

\subsection{Predicting Answer Correctness}
\label{section:features}
We train a simple feed-forward neural network (FFNN) with one hidden layer to predict whether the fine-tuned QA-model made an erroneous or correct answer span prediction for an input sequence $x_i$.
The FFNN is defined as follows,

\begin{equation}
    z_i = W_i^{M \times M}x_i^{M\times 1} + b_i^{M \times 1}
\end{equation}

\begin{equation}
    y_i = \sigma(W_i^{1 \times M}z_i^{M\times 1} + b_i^{1 \times 1}),
\end{equation}

where $\sigma$ denotes the sigmoid function.
The sigmoid function was applied to the FFNN's raw output logits since an answer span prediction could either be correct or incorrect.

\subsection{Model training}
\label{method:training}

\begin{table}[h!]
\centering
\begin{footnotesize}
\begin{tabularx}{\linewidth} {@{}XX|XX@{}}
\toprule
\multicolumn{2}{c}{\textsc{SQuAD}}&\multicolumn{2}{c}{\textsc{SubjQA}}\\
\midrule
\centering \textsc{Dev} & \textsc{Test} & \centering \textsc{Dev} & \textsc{Test} \\ 
\midrule
 \centering 700 & 843 & \centering 475 & 1145 \\
\bottomrule
\end{tabularx}
\end{footnotesize}
\caption{Number of examples in the leveraged development (i.e., train) and test sets.}
\label{tab:datasets}
\end{table}

Each FFNN is trained for a maximum number of 25 epochs until convergence. 
For optimization, we use Adam \cite{kingma2014adam} with a learning rate of $\eta = .01$ and a weight decay of .005 (this is equivalent to the L2 norm). Gradients are clipped whenever $||\frac{\partial_{L}}{\partial_{\theta}}|| \: \geq 10$. Input sequences are presented to the model in mini-batches of 8. The FFNN is implemented in PyTorch \cite{PyTorch}. Both BERT for QA and the FFNN are trained and evaluated on a single Titan X GPU with 12 GB memory. Usually, a dataset is split into three parts, namely a train, a development, and a test set, where the latter two splits together comprise approx. 20-30\% of the original dataset. Note that train and test datasets for SQuAD are equally large, and for SubjQA the test set even contains more than twice as many examples as the train set (see Table~\ref{tab:datasets}). Our train sets are the actual development sets (excluding unanswerable questions) with respect to the official QA datasets since we do not want to perform computations on hidden representations the QA model did produce during training. As such, we ascertain that the FFNN is trained on data the BERT model has never encountered during QA training. Hence, we cannot leverage development sets, and are limited to small training sets, which in turn further enhances the generalisability and potential of our approach.

We leverage one of the following three feature sets
as inputs to the FFNN,

\begin{enumerate}[noitemsep]
    \item \texttt{Raw}. For each sentence pair, $x_i$, we extract $\cos_{\tilde {H(a)}{_i}}$ and the standard deviation ($s$) w.r.t. the vector of cosine similarities among the rows of the matrix $\tilde {\mathbf{H}}_{a(i)} \in \mathbb{R}^{T_{a} \times P}$, where $i \neq j$, at every Transformer layer. Hence, $M = 2\times L$.
    
    \item \texttt{Approximation}. We multiply element-wise or concatenate $p_{(cdf)}^{l}$ $\forall \: l \in L$ with the \texttt{raw} cosine vector (see above). However, instead of knowing to which train distribution the observed $\cos_{\tilde H(a)_i}^{l}$ belongs, we approximate $p_{(cdf)}^{l}$ at test time with weighting $\cos_{\mathbf{\tilde H}(a)}^{l}$ w.r.t. correct and erroneous predictions differently (see Section \ref{method:approximating_cdfs}). Hence, $M=2\times L$ (weighting) or $M=2\times 2\times L$ (concat). 
    
    \item \texttt{CDF-aware}. Instead of approximating  $p_{(cdf)}^{l}$, the model is aware of whether an observed $\cos_{\tilde H(a)_i}^{l}$ must be interpolated given the distribution of correct or erroneous predictions. Again, the vector of $p_{(cdf)}^{l}$ $\forall \: l \in L$ is concatenated or multiplied element-wise with the \texttt{raw} cosine vector. Hence, $M=2\times 2\times L$ (concat) or $M=2\times L$ (weighting). This shows whether the FFNN benefits from information about the \textit{true} train CDFs, establishing the performance ceiling when approximating $p_{(cdf)}^{l}$ without information loss.
\end{enumerate}

\subsection{Approximating CDFs}
\label{method:approximating_cdfs}

Since at test time we are not aware of whether a BERT for QA model predicted an answer span correctly or erroneously, we have implemented two different weighting strategies to approximate the true $p_{(cdf)}^{l}$ $\forall \: l \in L$. $\cos_{\mathbf{\tilde H}(a)}^{l}$ denotes the train distribution of 
 $\cos_{\tilde H(a)_{i}}^{l}$ $\forall \: i \in N$ with respect to either erroneous or correct answer span predictions. Note, one must interpolate $\cos_{\tilde H(a)_{i}}^{l}$ given either of the two train distributions to yield $p_{(cdf)}^{l}$. Thus, one is required to infer to which of the two train distributions an observed $\cos_{\tilde H(a)_{i}}^{l}$ at test time probably belongs to. We approximated as follows,
 
\begin{enumerate}
    \item \texttt{Distance.} Here, we simply compute the distance between an observed $\cos_{\tilde H(a)_{i}}^{l}$ to the centroid of each of the two train distributions $\cos_{\mathbf{\tilde H}(a)}^{l}$. We leveraged the inverse distance as $w_{i}^{l}$, such as,
    
    \begin{dmath}
        {{w_{i}^{l}}_{(correct)} = 1 - (\cos_{\tilde H(a)_{i}}^{l} - {\mu_{i}^{l}}_{(correct)})} \\
        {{w_{i}^{l}}_{(incorrect)} = 1 - (\cos_{\tilde H(a)_{i}}^{l} - {\mu_{i}^{l}}_{(incorrect)})},
    \label{equation:mean_weighting}
    \end{dmath}
    
    where an $w_{i}^{l}$ is higher, if $\cos_{\tilde H(a)_{i}}^{l}$ happens to be closer to the mean of a train distribution.
    
    \item \texttt{CDF properties.} In this approximation, we exploited the mathematical properties of CDFs. In general, the smaller $|P(\cos_{\mathbf{\tilde H}(a)}^{l} \leq \cos_{\tilde H(a)_{i}}^{l}) - P(\cos_{\mathbf{\tilde H}(a)}^{l} \geq \cos_{\tilde H(a)_{i}}^{l})|$ is, the higher is the likelihood that an observed $\cos_{\tilde H(a)_{i}}^{l}$ belongs to this CDF as it denotes the area under the curve with the highest probability mass, which is considered the center. Hence, we exploited the inverse of the obtained value as $w_{i}^{l}$, such as,
    
    \begin{dmath}
        {{w_{i}^{l}}_{(correct)} =} \\ {1 - |P(\cos_{\mathbf{\tilde H}(a)}^{l}{_{(correct)}} \leq \cos_{\tilde{H(a)}_{i}}^{l})} \\ - {P(\cos_{\mathbf{\tilde H}(a)}^{l}{_{(correct)}} \geq \cos_{\tilde H(a)_{i}}^{l})|} \\
        {{w_{i}^{l}}_{(incorrect)} =} \\ {1 - |P(\cos_{\mathbf{\tilde H}(a)}^{l}{_{(incorrect)}} \leq \cos_{\tilde H(a)_{i}}^{l})} \\ - {P(\cos_{\mathbf{\tilde H}(a)}^{l}{_{(incorrect)}} \geq \cos_{\tilde H(a)_{i}}^{l})|},
    \label{equation:cdf_weighting}
    \end{dmath}
    
    where $w_{i}^{l}$ becomes large, the smaller $|P(\cos_{\mathbf{\tilde H}(a)}^{l} \leq \cos_{\tilde H(a)_{i}}^{l}) - P(\cos_{\mathbf{\tilde H}(a)}^{l} \geq \cos_{\tilde H(a)_{i}}^{l})|$ is.
\end{enumerate}

For both approaches, we approximated the true $p_{(cdf)}{_{i}^{l}}$ through a weighted sum of $p_{(cdf)}{_{i}^{l}}{_{(correct)}}$ and $p_{(cdf)}{_{i}^{l}}{_{(incorrect)}}$ through the following computation,

\begin{dmath}
    {p_{(cdf)}{_{i}^{l}} =} 
    \\ {\frac{1}{2} \times}
    {({p_{(cdf)}{_{i}^{l}}{_{(correct)}}} \times {w_{i}^{l}}{_{(correct)}})}\\
    {+ \: ({p_{(cdf)}{_{i}^{l}}{_{(correct)}}} \times {w_{i}^{l}}{_{(incorrect)}})}
\label{equation:approximation}
\end{dmath}

In initial experiments, we examined both approximation strategies. Due to \texttt{Distance} resulting in higher macro $F$1-scores than approximating through \texttt{CDF properties}, results are reported only for \texttt{Distance}. However, the difference between the two approaches was not significant, and might require further examination in follow-up studies.

\paragraph{Baselines}
\label{section:baselines}
We compare the features obtained with our method against the following three baselines: (i) \texttt{Majority}, (ii) \texttt{QA concat} (hidden representations of question and answer), and (iii) \texttt{Heuristic} (e.g.~n-gram overlap features).

\begin{enumerate}
    \item \texttt{Majority}. This approach simply predicts the most common class (i.e., correct or erroneous answer span prediction).
    \item \texttt{QA concat}. Concatenation of the average hidden representation w.r.t. the predicted answer span and question at last Transformer layer, where $x_{i} \in \mathbb{R}^{2 \times 768}$. Hence, $M = 1536$.
    \item \texttt{Heuristic}. Intuitively reasonable features, where $x_{i} \in \mathbb{R}^{9}$. Hence, $M = 9$. 
    \begin{enumerate}
        \item length of the predicted answer span;
        \item average $n$-gram overlap between predicted answer and question (i.e., BLEU score);
        \item cosine similarity between the average hidden representation w.r.t. the predicted answer span and question at last Transformer layer;
        \item vector of unigram, bigram, and trigram overlaps between predicted answer and question, normalized by the number of tokens in the answer and the question.
    \end{enumerate}
\end{enumerate}

\section{Results}
\label{section:results}
\subsection{Unsupervised QA Evaluation}

\begin{table}[t]
    \centering
        \fontsize{10}{10}\selectfont

    \begin{tabular}{lrr}
    \toprule
        \textsc{Method $\setminus$ Source} & \textsc{SQuAD} & \textsc{SubjQA} \\
        \midrule
        \textsc{majority} &  45.65\% & 42.11\% \\
        \textsc{qa concat} &  63.62\% & 46.44\% \\
        \textsc{heuristic}  &  87.23\% & 61.90\% \\
        \midrule
        \textsc{$\cos_{raw}$} &  49.19\% & 69.36\% \\
        \textsc{$\cos_{weight}$} &  63.63\% & 58.44\% \\
        \textsc{$\cos_{concat}$} &  55.28\% & 68.14\% \\
        \midrule
        \textsc{heuristic} $\oplus$ \textsc{$\cos_{raw}$} & 87.90\% & 74.56\% \\
        \textsc{heuristic} $\oplus$ \textsc{$\cos_{weight}$} & \textbf{88.43}\% & 72.29\% \\
        \textsc{heuristic} $\oplus$ \textsc{$\cos_{concat}$} & 88.33\% & \textbf{76.42}\% \\
        \midrule
        \textsc{$\cos_{weight}$} (\texttt{CDF-aware})&  78.58\% & 83.38\% \\
        \textsc{$\cos_{concat}$} (\texttt{CDF-aware})&  69.14\% & 90.46\% \\
        \bottomrule
    \end{tabular}
    \caption{Macro $F$1-scores for the binary classification task of predicting whether a fine-tuned BERT for QA model correctly or incorrectly predicted an answer span. $F$1-scores were averaged over five different random seeds. Best scores are depicted in bold face.}
    \label{tab:prediction_results}
\end{table}

Table~\ref{tab:prediction_results} shows that solely exploiting $\cos_{\mathbf{\tilde H}(a)}$ or additionally informing the model about $p_{(cdf)}$ outperforms all baselines for two out of three approaches when evaluating on SubjQA (rightmost column).
Interestingly, results slightly differ when examining QA-performance for SQuAD (centre column). 
The heuristics baseline yields a macro $F$1-score of 87.23\%, outperforming the two other baselines by a large margin, and performing better than our proposed approaches.
However, concatenating the features from the heuristics baseline with \texttt{raw} or \texttt{approximation}, further improves upon this strong baseline across both datasets, achieving 88.43\% and 76.42\% macro $F$1 for SQuAD and SubjQA respectively. 
For SubjQA, this leads to an absolute improvement of 14.5\% over the strongest baseline, and suggests that information about $\cos_{\mathbf{\tilde H}(a)}$ is decisive to predict a QA model's answer span prediction with respect to this dataset. 

The results obtained from the CDF-aware model show that further informing the model about $p_{(cdf)}$ $\forall \: l \in L$ has enormous potential to predict whether a BERT for QA model made a mistake or not.
The $F$1-score of 90.46\% indicates that mistakes might be predicted almost faultlessly by more sophisticated approximation methods, even without concatenating heuristic features.

Scaling $\cos{_{\tilde H(a)_{i}}}^{l}$ with $p_{(cdf)_{i}^{l}}$ appears to work better for SQuAD than concatenating $\cos_{\tilde H(a)_{i}}^{l}$ and $p_{{(cdf)}_{i}}^{l}$, whereas it is the other way around for SubjQA.
Hence, combining the two information sources is crucial across datasets but which merging strategy works best 
is dataset dependent, and might be a function of dataset complexity or number of context domains since these are the variables in which SubjQA and SQuAD differ.

\subsection{Error analysis}

We investigate two examples where \textsc{heuristic} features alone did not suffice to yield a correct prediction.
Given the question from SQuAD \textit{``What term did Eisenhower use to describe the character of communism?''}, the heuristic fails to identify the model's output as an incorrect answer.
Similarly, given the question from SubjQA \textit{``Are there any reviews on bath options at this hotel?''} and the correct model answer \textit{``great bathroom''}, the heuristic fails to identify this as a correct answer.
The concatenation of $\cos_{\tilde H(a)}$, and additional information about  $p_{(cdf)}$ were necessary to obtain correct predictions.
We further see that the observed $\cos_{\tilde H(a)}$ values are in line with our qualitative and statistical analyses. 
$\cos_{\tilde H(a)}$ is significantly higher in the final three Transformer layers for a correct prediction, and remains unchanged for erroneous predictions (see Table~\ref{tab:error_analysis} for more details).


\begin{table*}[tb!]
\centering
\begin{footnotesize}
\begin{tabularx}{\linewidth} {@{}l||X|X@{}}
\toprule
\textsc{Source}&\multicolumn{1}{c}{\textsc{SQuAD}}&\multicolumn{1}{c}{\textsc{SubjQA}}\\
\midrule
\midrule
\centering \texttt{Question}  & \centering "What term did Eisenhower use to describe the character of communism?" & \centering "Are there any reviews on bath options at this hotel?" \cr \\
\centering \texttt{Answer}  & \centering "atheistic"  & \centering "great bathroom" \cr  \\
\midrule
\texttt{BERT QA}  & \centering \texttt{incorrect} & \centering \texttt{correct} \cr \\
\midrule
\textsc{heuristic}  & \centering \texttt{correct} \xmark & \centering \texttt{incorrect} \xmark \cr  \\
\textsc{heuristic} $\oplus$ \textsc{$\cos_{w}$}  & \centering \texttt{incorrect} \cmark  & \centering \_ \cr \\
\textsc{heuristic} $\oplus$ \textsc{$\cos_{c}$}  & \centering \_  & \centering \texttt{correct} \cmark \cr \\
\midrule
\centering $\cos_{\tilde H(a)}^{l}$ $\forall l \in L$ & \centering $[0.07, 0.16,  0.26, 0.27, 0.31, 0.20]$ &  \centering $[0.06, 0.14, 0.29, 0.62, 0.63, 0.55]$ \cr \\
\bottomrule
\end{tabularx}
\end{footnotesize}
\caption{Error analysis. $\cos_{\tilde H(a)}^{l}$ is presented for each of the six Transformer layers.}
\label{tab:error_analysis}
\end{table*}

\section{Related Work}
\label{section:related_work}
Since automatic evaluation is considered an important topic in other areas of NLP, e.g. MT \citep{DBLP:conf/acl/PapineniRWZ02} and summarisation \citep{DBLP:conf/naacl/OwczarzakCDN12}, we want to draw attention to such techniques for QA. To the best of our knowledge, ours is the first proposal unsupervised QA evaluation method.

One recent study which we take inspiration from present a layer-wise analysis of BERT's Transformer layers to investigate how BERT answers questions \cite{bert_qa_layerwise}. For each Transformer layer, they project the model's hidden representations into $\mathbb{R}^{2}$ to illustrate how BERT clusters different parts of an input sequence 
while searching for an answer span. We replicate their findings for SQuAD, and show that this insight holds across two datasets and seven domains through observing the same patterns w.r.t. SubjQA. However, we use this qualitative analysis just as an initial step from which we start extracting information to develop an unsupervised QA evaluation method.

\citet{DBLP:journals/corr/abs-1910-06431} investigate which tokens in sentence pairs receive particular attention by BERT's self-attention mechanisms to answer a question, and how the multi-headed attention weights change across the different layers. Similarly to \citet{bert_qa_layerwise}, the authors did solely conduct a qualitative analysis of the model. Contrary to \citet{bert_qa_layerwise}, the study focuses on a single implementation of BERT and exclusively exploited SQuAD \cite{DBLP:journals/corr/RajpurkarZLL16,DBLP:journals/corr/abs-1806-03822} without inspecting BERT's behaviour with respect to other, more challenging QA datasets where contexts belong to different domains. The latter is particularly important for real-world settings, which is why we also evaluate on SubjQA. 

\section{Discussion}
\label{section:discussion}


The heuristic method we investigate in this work, based on features such as n-gram overlap between question and answer, yielded surprisingly high results on SQuAD.
On the other hand, the results for SubjQA were quite low when using the heuristic only.
This shows that, although a simple heuristic might be sufficient for a single dataset, it does not necessarily generalise across datasets and domains.

Conversely, our proposed method, which takes answer span similarities into account, was highly successful on SubjQA without the need for any heuristic features, but only outperformed the SQuAD baseline by 15-20\% macro F1 score.
Combining the two methods yielded the best results across both data sets and all domains.
This demonstrates that the information contained in the heuristic approach and in our proposed method are complementary.

\subsection{Error analysis}
\label{discussion:error_analysis}
The concatenation of $\cos_{\tilde H(a)}$, and further information about  $p_{(cdf)}$ were necessary to obtain correct predictions.
The observed $\cos_{\tilde H(a)}$ values are in line with our qualitative and statistical analyses. 
$\cos_{\tilde H(a)}$ is significantly higher in the final three Transformer layers for a correct prediction, and remains unchanged for erroneous predictions (see last row in Table~\ref{tab:error_analysis}). It is interesting to note that for a correct answer span prediction $\cos_{\tilde H(a)}$ increases considerably from layer 3 to layer 4, but no notable change can be observed thereafter.

\section{Conclusion}
\label{section:conclusion}
We have shown that the hidden representations of answers in transformer-based models can be used to predict whether or not that answer is correct.
In combination with heuristic methods, we are able to predict the correctness of answers with a macro F1 score of 88.38\% for SQuAD and 76.42\% for SubjQA.
Apart from the applications in unsupervised evaluation of QA, we expect that this method can be applied to semi-automatic generation of QA datasets.

\section*{Acknowledgements}
The authors would like to thank the anonymous reviewers for their feedback which  contributed to improving the final version of the paper.

\bibliographystyle{acl_natbib}
\bibliography{anthology,references}

\end{document}